\documentclass{article}

\usepackage{arxiv}

\usepackage[utf8]{inputenc} 
\usepackage[T1]{fontenc}    
\usepackage{hyperref}       
\usepackage{url}            
\usepackage{booktabs}       
\usepackage{amsfonts}       
\usepackage{nicefrac}       
\usepackage{microtype}      
\usepackage{natbib}
\usepackage{graphicx}
\usepackage[
  separate-uncertainty = true,
  multi-part-units = repeat
]{siunitx}
\usepackage{subcaption}
\usepackage{csvsimple}
\usepackage{amsmath}

\usepackage[capitalize]{cleveref}

\title{Balancing Accuracy and Latency in Multipath Neural Networks}

\author{
  Mohammed Amer\thanks{Corresponding author} \\
  School of Computer Science\\
  University of Nottingham Malaysia\\
  Semenyih, Malaysia \\
  \texttt{hcxma1@nottingham.edu.my} \\
   \And
  Tom\'as Maul\\
  School of Computer Science\\
  University of Nottingham Malaysia\\
  Semenyih, Malaysia \\
  \texttt{tomas.maul@nottingham.edu.my} \\
  \And
  Iman Yi Liao\\
  School of Computer Science\\
  University of Nottingham Malaysia\\
  Semenyih, Malaysia \\
  \texttt{iman.liao@nottingham.edu.my} \\
}

\begin{document}

\maketitle

\setcounter{footnote}{0}

\begin{abstract}

The growing capacity of neural networks has strongly contributed to their success at complex machine learning tasks and the computational demand of such large models has, in turn, stimulated a significant improvement in the hardware necessary to accelerate their computations. However, models with high latency aren't suitable for limited-resource environments such as hand-held and IoT devices. Hence, many deep learning techniques aim to address this problem by developing models with reasonable accuracy without violating the limited-resource constraint. In this work, we use a one-shot neural architecture search model to implicitly evaluate the performance of an intractable number of multipath neural networks. Combining this architecture search with a pruning technique and architecture sample evaluation, we can model the relation between the accuracy and the latency of a spectrum of models with graded complexity. We show that our method can accurately model the relative performance between models with different latencies and predict the performance of unseen models with good precision across different datasets. 

\end{abstract}

\section{Introduction}

Deep learning has gained momentum as a machine learning approach that has surpassed classical techniques in a broad category of complex tasks by introducing multiple layers of feature composition. As a side effect, the computational complexity of its models grew substantially and they are becoming exceedingly demanding with the accelerating research in the field. The field has observed a strong correlation between increasing model sizes, in a coordinated way, and the accuracy gain in different tasks. Applications that utilize a compute-centric approach, i.e central repository of accumulated data and high computational resources that can be used for training and enhancing models, could scale their computational resources and data size sufficiently to accommodate for the exploding demand of the bigger-size data-greedy models.

However, not every application lends itself to central computation. Hand-held devices and IoT (Internet of things) are pervasive technologies with rich data. For many concerns like privacy, security and bandwidth, these data can't be centralized and a different approach is needed. The data-centric approach, hence, depends on distributed computing and learning across many terminal devices that are generally limited in resources.

Latency refers to the increased processing runtime of a neural network, usually as measured at inference time. While there are implementation and hardware factors that can contribute to a network's latency and cause the same architecture's latency to vary across devices \citep{Yang2020}, the parameter count of a network is a very important proxy measure of a network's latency, specially when benchmarking similar architectures \citep{Howard2017,Sandler2018,Iandola2016,Hinton2015}. Also, since the parameter count offers a good proxy measure of the model's memory complexity, it is considered an overall good measure of the suitability of a model to limited-resource applications. In this work, we similarly rely on the parameter count as a proxy of the model latency at inference time.

Since deep learning is computationally demanding, ways of reducing its complexity, while maintaining its accuracy are needed to harness this potential. Several methods were researched in the quest to achieve this goal. Separable convolutions were used in a set of techniques \citep{Howard2017, Sandler2018, Iandola2016} to reduce network size and complexity. Quantization, hashing and pruning techniques \citep{Han2015} aim at reducing parameter precision, and allow sharing of parameters based on activations and sparsifying parameters, respectively. Factorization \citep{Jaderberg2014} reduces the number of parameters by approximating complex computations by another smaller set of factored computations. Distillation \citep{Hinton2015} effectively compresses knowledge from a large network into another smaller network.

Neural Architectural Search (NAS) is a set of related techniques for optimizing neural architectures by searching in a prespecified search space. Vanilla NAS techniques depend on sampling architectures from the search space. These architectural samples are then trained and evaluated in order to guide the search process towards more promising architectures. This sampling process can happen through techniques like Reinforcement Learning (RL) \citep{Zoph2016,Zoph2018}, Evolutionary Algorithms (EAs) \citep{Liu2017} and random search \citep{Liu2017,Zoph2018}.

The training and evaluation of a large number of sampled architectures are computationally expensive processes that need a serious scale of distributed computing. One-shot NAS techniques \citep{Liu2018,Zhao2020} were developed to avoid these heavy computational needs. A supernet is a large network that spans multiple subnetwork architectures with overlapped weight sharing. By single-time training of a supernet that is designed to span some search space, one-shot NAS can approximate the evaluation of an exponential number of architectures without the need for exhaustive training and evaluation of individual networks. Specialized NAS techniques were developed with search spaces aimed at limited resource conditions \citep{Tan2019, Tan2019a, Cai2018}. 

MpathNNs (Multipath neural networks) are NNs that have multiple, usually independent, paths. Compared to another model having the same width and depth, MpathNNs have a sparser set of parameters. MpathNNs have been used in several previous studies for improving generalization \citep{KienTuongPhan2015,Ciresan2012,Wang2015}, image captioning \citep{Yu2019}, feature extraction \citep{Yu2019a}, cross modal learning \citep{Hong2015, Hong2019}, and dimensionality reduction \citep{Zhang2018b}.

We can think of MpathNNs as a way of sparsifying a fully connected (FC) model with similar width and depth (in general, the same architectural hyperparameters excluding the number of learned parameters). However, it isn't clear how to divide such an FC model into multiple paths specially given that the number of possible MpathNNs corresponding to an FC model is intractable to be evaluated thoroughly. We utilize a one-shot model to evaluate the relative importance between such divisions. We can, then, prune the network using this information to get an MpathNN with decent balance between latency and accuracy. We further construct an approximate relation between complexity and accuracy by fine-tuning and validating a small sample of possible models. This allows a modeler to satisfy a given accuracy-latency requirement ahead of any specific MpathNN model training. Finally, the model can be fine-tuned for a given complexity to yield the final desired model. Our contributions are:

\begin{itemize}
    \item Using a one-shot model to approximate the relative accuracy across all possible divisions of paths of an FC model.
    \item We utilize the information gained from the search process to get MpathNNs models with a desired path count through a pruning process.
    \item We use a predictive model learned from a small architectural sample to predict potential model accuracy ahead of any specific model training/fine-tuning. 
\end{itemize}

\section{Literature Review}

Biological nervous systems exhibit multipath branching and parallel computations in their structure \citep{Gollisch2010, Otsuna2014}. This inspired Artificial Neural Networks (ANNs) that adopt similar structures. Multipath Convolutional Neural Network (CNN) was used by \citet{Ciresan2012} where different paths have a different preprocessed version of the same input. The outputs of different columns are averaged to produce the final output. In a similar approach by \citet{Wang2015}, different paths are fed with differently filtered versions of the input and the output is consolidated from the multipaths by a fully connected subnetwork. In image captioning, \citet{Yu2019} use several detectors applied in parallel for multi-view feature learning. For unsupervised dimensionality reduction, \citet{Zhang2018b} use the same idea by applying local contractive autoencoders to extract local features and then apply affine transformations to align with a global coordinate system. \citet{Yu2019a} extract hierarchical features of different granularity using a parallel structure applied to both image and text data. For human pose recovery (HPR), \citet{Hong2015} extract 2D and 3D features from different modalities and learn to map 2D to 3D features. Similarly for HPR, \citet{Hong2019} feed inputs from different modalities into different branches, each with its own reconstruction task.

Inception-v1, the ANN that won ILSVRC-14, was introduced by \citet{Szegedy2015a} and it exhibits highly branched structure. Inception-v2/v3 \citep{Szegedy2015b} and Xception \citep{Chollet2016} improved over inception-v1 by exploiting more multibranching. ResNetXt \citep{Xie2016} and Residual Inception \citep{Zhang2018a} are variants of ResNet \citep{He2016} where the modular block has multipath structure. FractalNet, proposed by \citet{Larsson2016}, has a recursive highly branched structure. The parallel circuit network by \citet{Phan2016} is based on a multipath structure and showed generalization improvement over fully connected (FC) networks when regularized by DropCircuit \citep{Phan2018}. PathNet \citep{Fernando2017} is a super NN (a large NN that is supposed to contain multiple subnetworks) that is targeted at learning sequential tasks efficiently. The network is highly branched and a genetic algorithm is used to select which subnetwork to learn. After the convergence of each task, the fittest path is frozen before moving to the next task.

Due to the explosion of NN sizes and the need to run on limited resource devices, several techniques have been investigated to reduce network sizes while maintaining as much accuracy as possible. Several techniques have been researched, including, utilization of separable convolutions, quantization, hashing, pruning, factorization, Neural Architectural Search (NAS) and distillation. \citet{Howard2017} used separable convolutions and 1x1 convolutions in MobileNet to reduce complexity. More fine control over architecture size was done by width and resolution multipliers, two hyperparameters that control network width and representation resolution, respectively. MobileNetv2 \citep{Sandler2018} enhanced over MobileNet by introducing bottlenecks and inverted residuals. A similar approach was used in \citet{Iandola2016} by introducing two different types of cells, squeeze and expansions cells, containing 1x1 filters or a mix of 1x1 and 3x3 filters, respectively. \citet{Chen2015} used the hashing trick \citep{Weinberger2009, Shi2009} to share weights based on feature activation. \citet{Han2015} used a combination of pruning, quantization/Huffman-coding and fine tuning. After training, the network is pruned based on the weight magnitudes. The sparse weights are then quantized and Huffman-coded and, finally, fine-tuned. The idea by \citet{Jaderberg2014} was to factor convolutional filters into rank-1 horizontal and vertical vectors. The separation of the kernels was based on gradient-guided optimization. Knowledge distillation \citep{Hinton2015} is a knowledge transfer technique that trains a smaller student network using soft targets generated from a larger teacher network. The soft targets are generated from the teacher using a softmax with high-temperature. Two objectives are used for the training, the first is based on the cross entropy of the soft targets with the same high-temperature used for generating the soft targets and the other is based on the actual labels with a softmax temperature of 1.

Neural Architectural Search (NAS) is the set of techniques targeted towards the automated optimization of ANN architectures. \citet{Zoph2016} use an RNN to sample an ANN architecture. The sampled architecture is then trained and evaluated to provide the necessary reward that will be used to train the RNN through RL. \citet{Zoph2018} use a similar RL technique, however, instead of searching for a complete ANN architecture, they restrict their search to finding a cell/block that will be repeated later to build a full architecture. \citet{Zhang2020} propose a NAS technique for DenseNet \citep{Huang2016a}. The core idea is to prune unnecessary skip connections by formulating the pruning process as a Markov Decision Process (MDP). An RL agent is then trained to suggest the potential connections for pruning.

As discussed, one-shot NAS depends on training a supernet in order to implicitly evaluate the performance of many architectures in parallel without the need for exhaustive evaluation of individual candidates. \citet{Saxena2016} use a supernet that spans a CNN search space by arranging the network architectural hyperparameters across three axes: (1) a layer axis, which is analogous to the usual depth axis, (2) a scale axis, which spans feature maps with different resolutions and (3) a channel axis, across which the number of channels vary. DARTS \citep{Liu2018} is another differential NAS technique that uses a set of stacked modular multipath cells, each having a branched structure with different operations. The operations are combined by a parameterized weighting function and the whole model is optimized end-to-end for architectures with higher accuracy. PC-DARTS \citep{Xu2019} mitigates the high memory need of DARTS by downsampling the number of channels of each node input. To avoid fluctuations in the architecture optimization due to this random sampling process, node outputs are combined in a normalized learnable way, instead of concatenation like in DARTS. A technique similar to DARTS is used by \citet{Cai2019} where the weighting factors of the different operations are produced by a secondary network. A highly branched network is used in a NAS technique by \citet{Bender2018} to evaluate multiple branched NNs in parallel. Random paths are dropped during training as a form of regularization and subnetworks are evaluated by sampling from the trained supernet.

Some NAS techniques were developed with the aim of searching architectural spaces that suit limited resource applications in mind. \citet{Tan2019} used a search space based on diverse, instead of homogeneous, blocks. The architectural search optimizes for accuracy subject to a constraint of maximum allowed latency. The latency is measured as the runtime on actual mobile devices, instead of FLOPS. \citet{Tan2019a} analysed the effect of scaling the width, depth and resolution of the network, and based on that, modified the work by \citet{Tan2019} to enforce uniform scaling. \citet{Cai2018} used a NAS technique that samples different potential operations from the search space, in contrast to techniques like DARTS \citep{Liu2018} that use a linear mixture of possible operations.

The approach most similar to our technique is slimmable NNs \citep{Yu2018a}. Slimmable NNs aim to balance network complexity and accuracy by activating different fractions of the total network width. The network is trained by accumulating gradients from a predefined set of width fractions.

In this work we use a one-shot model to evaluate different possible divisions of a fully connected network into independent paths and, then, guide the pruning process to get potential MpathNN models. Our work is different from \citet{Yu2018a} in that the whole width of the network is active since we aim at evaluating all possible divisions. The one-shot model, in contrast to previous work, is used to draw a relation between parameter count and validation error, which can aid the modeler to realize the required accuracy-latency balance beforehand, without the need for training multiple models from scratch.

\section{Methodology}


We define a Multipath Neural Network (MpathNN) as a Neural Network (NN) having two or more independent paths  (i.e having no interconnections). We also define an MpathNN-equivalent MLP (Multilayer Perceptron), which we will refer to by an equivalent MLP for convenience, as an NN having the same number of layers (i.e depth) and number of nodes per layer (i.e width). In the same way, we will call any MpathNN that has an equivalence to a given MLP as a derived MpathNN. An equivalent MLP has certainly more capacity in terms of its parameter count than any corresponding derived MpathNN. Pruning an MLP into a number of paths is done by dropping a set of connections in the way described later. Thus, a derived MpathNN is more computationally and memory efficient, at the expense of losing capacity, which may lead to accuracy degradation. For some applications, like models targeted at devices with limited resources, some degree of accuracy degradation can be tolerated in order to achieve a more efficient runtime and less memory consumption.

\begin{figure}[h]
    \centering
    \includegraphics[width=0.5\textwidth]{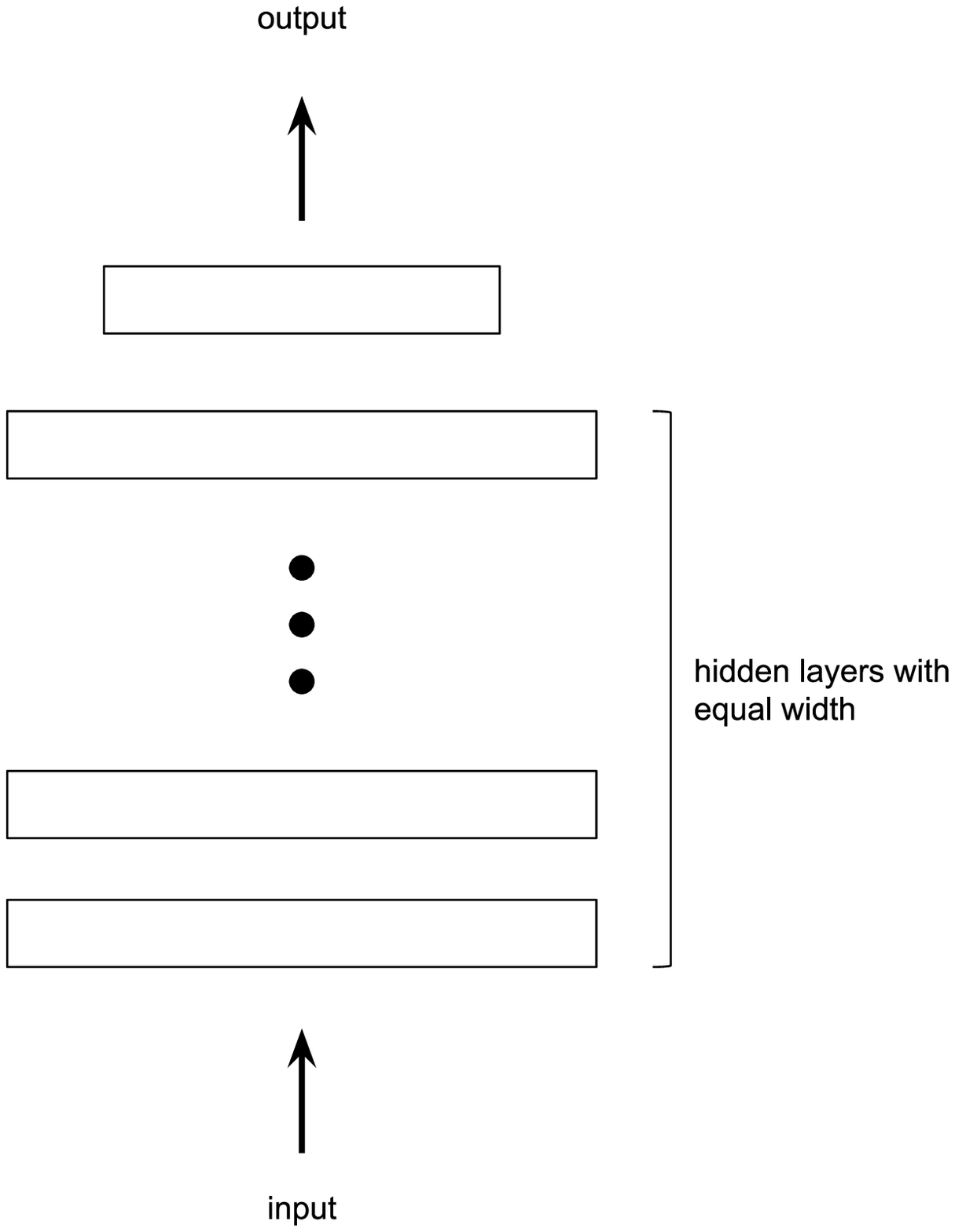}
    \caption{Simplified Equivalent MLP Architecture}
    \label{fig:mpathnn-eq-mlp}
\end{figure}

Due to the limited resources in these kinds of applications, usually a model is trained on more powerful machines, and used later for inference on the resource-constrained device. At one extreme, one can imagine training every possible derived model and choosing the one with the optimal balance between accuracy and complexity. This is, however, not practically possible for almost any realistic MLP due to a combinatorial explosion\footnote{Calculating the number of unique derived MpathNN models isn't a trivial problem and is beyond conventional combinatorial techniques and the scope of this work. The problem is related to the integer partitions problem which can be solved using generating functions or through an approximation formula \citep{Andrews1998,Andrews2004}. In a more recent work, \citet{Bruinier2013} found a closed-form formula which is, nonetheless, not trivial. For our purposes, it is sufficient to state that for a model following the simplified scheme shown in \cref{fig:mpathnn-eq-mlp} with width $W$, the number of unique architectures grows sub-exponentially in $W$ \citep{Andrews2004} and that for widths as small as 64 and 128, the number of unique architectures is on the order of 1.7M and 4.4B, respectively.}.

\begin{figure}[h]
    \centering
    \includegraphics[width=\textwidth]{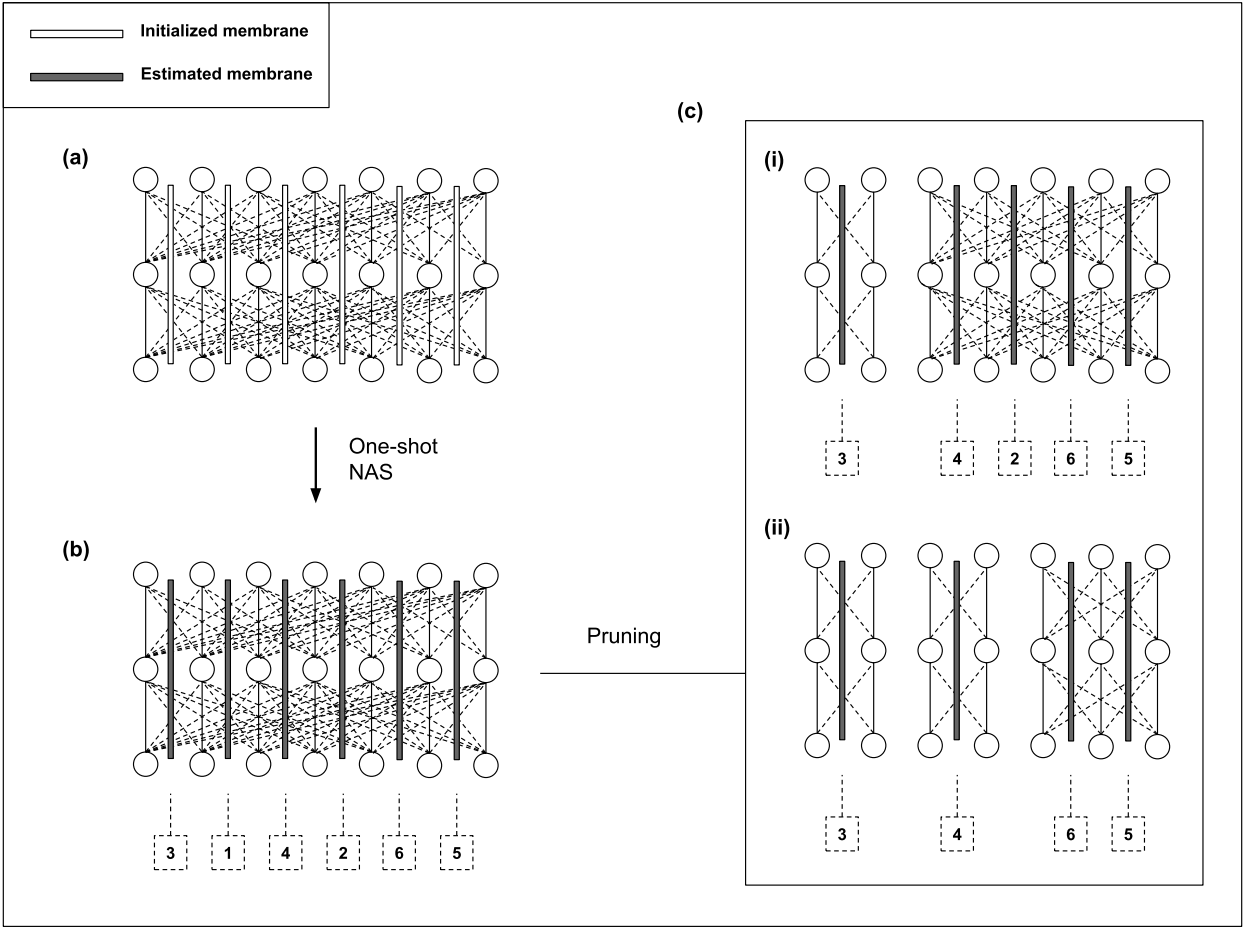}
    \caption{One-shot NAS of an MpathNN with 3 hidden layers and a width of 7 nodes. The input connections and the output layer were omitted for clarity. (a) An equivalent MLP is turned into a one-shot model by inserting a set of randomly initialized membranes. (b) The estimated membrane permeabilities are ranked in ascending order. (c) Pruning can be used to get an MpathNN with a given path count, e.g MpathNN with 2 paths (i) or 3 paths (ii).}
    \label{fig:balance-oneshot}
\end{figure}

Our aim in this work, is to use a method similar to one-shot models \citep{Liu2018, Xu2019, Cai2019, Bender2018, Yu2018a, Saxena2016} to approximate this kind of accuracy-latency relation, without the need to train every possible derived model. \cref{fig:balance-oneshot} shows a sketch of the one-shot NAS and the pruning steps of our method. Our method consists of two phases. First, a modified equivalent MLP model is fully-trained to estimate the relative importance of different path divisions. Second, a sample of derived models generated from the trained model by pruning are fine-tuned for a small number of epochs and, then, validated to establish a relation between capacity and validation accuracy. Based on this relation, a modeler can select the model with the desired balance between accuracy and complexity and fine-tune it to get the final inference-ready model.   

For simplicity of analysis and efficiency of implementation purposes, we consider only MpathNNs which follow the structure of the equivalent MLP depicted in \cref{fig:mpathnn-eq-mlp}. The set of models following this general architecture consist of a number of hidden layers, all having the same width (i.e number of nodes). The first hidden layer (i.e the layer receiving the input) is a Fully Connected (FC) layer. The remaining upper hidden layers are divided into a number of independent paths, each having the same width for all of their layers. The output layer is an FC layer much like the first hidden layer, receiving all path outputs as a concatenated vector. 

\begin{figure}[h]
    \centering
    \includegraphics[width=.6\textwidth]{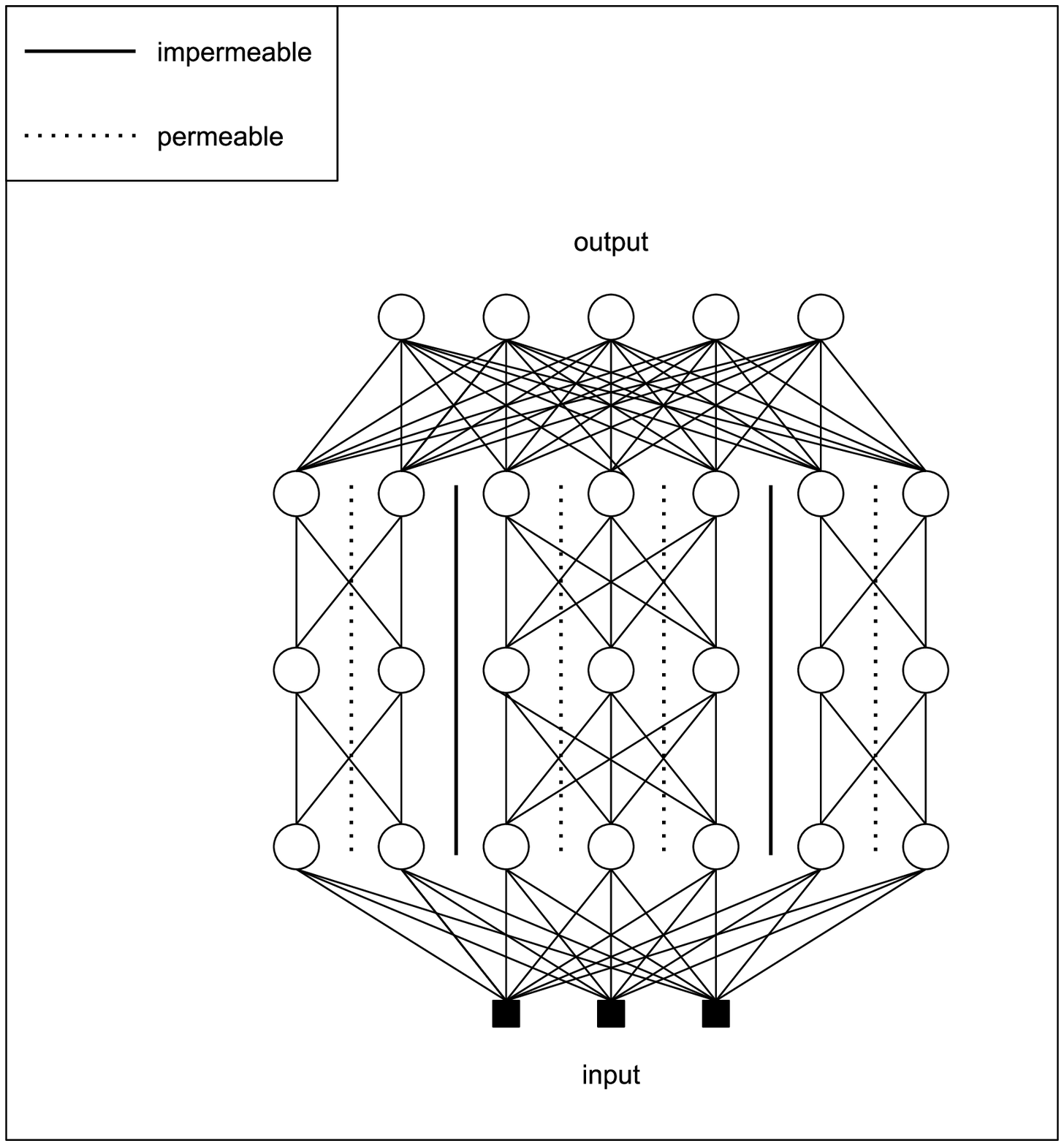}
    \caption{MpathNN Example with 3 paths}
    \label{fig:mpathnn}
\end{figure}

We can think of an MpathNN as a variation of an FC network, where we draw an imaginary line, which we will call a membrane, between each two adjacent neurons across all the hidden layers starting from the second hidden layer up to the final hidden layer. This membrane has a permeability that ranges from impermeable ($0$) to fully permeable ($1$). We, then, allow connections based on this permeability, with full connection strength if the membrane is fully permeable, no connection if it is impermeable and a graded connection strength in between. This means that for an MpathNN of width $W$, if an impermeable membrane is located between the two adjacent neurons $n_i$ and $n_{i+1}$ with indices $i$ and $i+1$, then for any two successive hidden layers, no connections going into neuron $n_{i+1}$ from its previous layer are allowed from the set of neurons $\{n_k \mid k \leq i\}$ and, similarly, no connections are allowed into a neuron $n_i$ from the set of neurons $\{n_l \mid l \geq i+1\}$. An example diagram is shown in \cref{fig:mpathnn}.

We think of each of the described membranes between any two adjacent neurons as having a permeability $p(m_i) \in [0,1]$ described by

\begin{equation}\label{eq:perm-special}
    p(m_i \mid \Omega_i) = \sigma(\Omega_i) \mid \forall m_i \in \mathbb{M}
\end{equation}

where $\mathbb{M}$ is the set of membranes with cardinality $|\mathbb{M}|$ equal to $W-1$ (where $W$ is the network width), $m_i \in \mathbb{M}$ is the membrane between neurons $n_i$ and $n_{i+1}$, $\sigma$ is the sigmoid nonlinearity and $\Omega$ is the set of weights parameterizing the membrane permeabilities. Note that $\Omega$ is shared between all hidden layers. Also, note that at inference time, we need to derive an MpathNN, hence, a membrane is either permeable, so that it adds no paths to the architecture, or impermeable and so it contributes one additional path to the modular architecture. However, in order to perform a differential search on our one-shot model, we need the permeability function to be differentiable. We will discuss later how we use pruning to discretize these permeabilities to obtain an MpathNN.

As we have a set of membranes spanning the whole network between each two adjacent neurons, we need to model the general connectivity between any two neurons with connections crossing multiple membranes. In order to model a connection between any two neurons $n_i$ and $n_j$, we generalize equation \cref{eq:perm-special} by considering the set of membranes that will be crossed by the given connection to connect the two neurons

\begin{equation}
    p(c_{i \rightarrow j} \mid \Omega_{i:j-1}) = \min_{k \in [i, j-1]} \sigma(\Omega_k)
\end{equation}

where $c_{i \rightarrow j}$ is the connection from neuron $n_i$ in the hidden layer $l-1$ into neuron $n_j$ in the following hidden layer $l$ and $i < j$. Note that due to symmetry, $p(c_{j \rightarrow i}) = p(c_{i \rightarrow j})$. In other words, when a connection crosses multiple membranes, we model it by the lowest membrane permeability it crosses. The reasoning is that a single impermeable membrane is sufficient to hinder a connection.

To create our one-shot model, we need to integrate these permeabilities into our equivalent MLP feedforward pass so that we can perform an end-to-end optimization. We do this by multiplying each weight by its corresponding permeability in the feedforward pass. In general,

\begin{equation} \label{eq:balance-oneshot}
    \widehat{w}^{(l)}_{i,j} = w^{(l)}_{i,j} * p(c_{i \rightarrow j})
\end{equation}

where $w^{(l)}_{i,j}$ is the weight connecting neuron $n_i$ in layer $l-1$ to neuron $n_j$ in layer $l$. The resulting modified weight matrix is then used for carrying on the feedforward computation in the usual way. We, then, train the model parameters (including membranes parameters) till convergence. We refer to this phase as the one-shot training. 

After the model convergence, we will have the estimations of the membrane permeabilities, with each permeability $p(m_i) \in [0,1]$. To have an MpathNN with a given path count from the converged model, we need a way to select a set of membranes to prune (i.e to make impermeable), based on their corresponding permeabilities. We note that adding one more impermeable membrane to an equivalent MLP will add one more path to the resulting MpathNN. This means that the range of possible path counts $M$ in a derived MpathNN with width $W$ is $[2,W]$. For each possible path count, there will be many possible ways to make the additional division (i.e to place the additional impermeable membrane) and each division will give rise to an MpathNN with a different capacity (i.e parameter count), and, hence, with a different accuracy in general. To choose a good division, we follow the following pruning technique. We select which membranes to make impermeable by ordering the membrane permeabilities in ascending order. Then, to have an MpathNN with a given path count $b$, we set to zero the ordered membrane permeabilties with indices in the range $[1,b)$. This is motivated by the intuition that weights multiplied by a small permeability will be less important for the MpathNN accuracy. We set the permeabilities of the remaining permeable membranes to one. 

While each path count can have many corresponding MpathNNs with different parameter counts, under the described pruning technique, each path count will be mapped to a specific MpathNN with a unique parameter count. To establish a relation between the path count, and hence the parameter count, and accuracy, we need to obtain the validation accuracies for a sample of MpathNNs with path counts that sufficiently cover our search space. Given some sample size, we choose to sample a set of path counts $\mathbb{B} = \{b: b \in [2,W]\}$ that are evenly spaced from each other over the specified interval $[2,W]$. For example, for $W = 10$ and a sample size of five, we choose $\mathbb{B} = \{2, 4, 6, 8, 10\}$. Then, for each path count $b \in \mathbb{B}$, we obtain the corresponding MpathNN with $b$ number of paths from the one-shot model using the described pruning strategy. We, then, fine-tune each of these MpathNNs for a small number of epochs and calculate their validation accuracies. We refer to this phase as sample fine-tuning.

At the end, we are left with a set of path counts $\mathbb{B}$ and their corresponding validation accuracies $V$. As we discussed, path counts can be uniquely mapped to parameter counts under the defined pruning technique. Despite the fact that we are ultimately interested in the parameter counts, we use the path counts as the input to the regression model in order to make the regression more stable. The reason is that path counts are a well-behaving arithmetic sequence of integers, while their corresponding parameter counts are a sequence of integers with irregular gaps. We, hence, fit a simple linear regression model using path counts as an input and the corresponding validation accuracies as the target. To do that, we first order the path counts in an ascending order and reorder the corresponding accuracies $V$ to maintain the alignment between each path count and its associated accuracy. With some abuse of notation, we will refer to this ordered array also by $\mathbb{B}$. Then, an ordinary least-squares regression model is trained to predict the accuracy array from the path count array. The regression model is, then, a map $g: \mathbb{N}^{|\mathbb{B}|} \rightarrow [0,1]^{|\mathbb{B}|}$ with input and output sizes $|\mathbb{B}|$ and the training data is just a single point consisting of the array pair $(\mathbb{B}, V)$. The reason for not treating the path counts and their associated accuracies as multiple data points of a scalar dimensionality is that the mapping of the accuracy would be reduced to a linear function of the path count. This would limit the prediction accuracy significantly. While fitting the model to the path counts and the accuracies as a single point in a higher dimensional space is still a linear transformation, this will increase the prediction accuracy by utilizing more information from the relation between single values. Alternate methods, like polynomial fitting or non-linear models, can be used to achieve a similar result, however, at the cost of introducing more complexity or additional hyperparameters.

We use this regression model to predict the validation accuracy for any path count not in the regression data by the following method. We insert the path count that we need to predict the accuracy of in the ordered path count array $\mathbb{B}$ in a correctly ordered position. For example, if our path count array is $\mathbb{B} = \{2, 4, 6, 8\}$ and we want to predict the accuracy of MpathNN with $5$ paths, then we have $\mathbb{B}^{\prime} = \{2, 4, 5, 6, 8\}$ where we have inserted $5$ at index $3$ (indices start from $1$). To maintain the correct input size of our regression model (4 for this example), we need to remove one of the old elements in $\mathbb{B}^{\prime}$. We do that by removing the first element of the ordered array if the new path count was inserted as the last element or by removing the last element if it was inserted anywhere else. For our example, since the new path count $5$ wasn't inserted as the last element, we remove the last element in the new array which becomes $\mathbb{B}^{\prime} = \{2, 4, 5, 6\}$. After that, we can use our trained regression model to map the new array $\mathbb{B}^{\prime}$ into a predicted accuracy array $V^{\prime}$. The accuracy prediction for our new path count is then acquired from the same index in $V^{\prime}$ (i.e the insertion index used in $\mathbb{B}^{\prime}$), which for our example would be index $3$.

Using the described inference process, a modeler, having a range of required accuracies and an upper limit on resources, can make the required trade off between the two quantities by estimating the validation accuracy, and hence an approximate test accuracy, for any path count and its corresponding parameter count, without the need to train the intractable number of all possible MpathNN models. As we discussed, while we use the path counts for regression, they are uniquely convertible to parameter counts since the pruning process produces a unique MpathNN for each path count.

\section{Experiments} \label{sec:membrane-exp}

To assess our method, we test three different architectures, following the simplified general architecture described earlier, on three different datasets. We will refer to each architecture by the abbreviation [mlp or mpath]-[hidden layers number]-[hidden features]. The prefix is mlp if it is an equivalent MLP or mpath if it is an MpathNN model. We follow the prefix by the number of hidden layers and, then, the number of hidden features in each hidden layer. Our three architectures have 4, 4 and 6 layers and 500, 200 and 200 hidden features, respectively. For each of the used architectures, we train an equivalent MLP to act as a baseline. 

\begin{table}[h]
    \centering
    \begin{tabular}{c|c|c|c}
         Model & \# hidden layers & \# hidden features & Test error (\%)\\
         \hline
         mlp-4-500 & 4 & 500 & 1.49\SI{\pm 0.06}{}\\
         \hline
         mpath-4-500 & 4 & 500 & 1.64\SI{\pm 0.05}{}\\
         \hline
         mlp-4-200 & 4 & 200 & 1.79\SI{\pm 0.1}{}\\
         \hline
         mpath-4-200 & 4 & 200 & 1.87\SI{\pm 0.05}{}\\
         \hline
         mlp-6-200 & 6 & 200 & 1.7\SI{\pm 0.1}{}\\
         \hline
         mpath-6-200 & 6 & 200 & 1.82\SI{\pm 0.1}{}\\
         \hline
    \end{tabular}
    \caption{MNIST results}
    \label{tab:balance-mnist-results}
\end{table}

\begin{table}[h]
    \centering
    \begin{tabular}{c|c|c|c}
         Model & \# hidden layers & \# hidden features & Test error (\%)\\
         \hline
         mlp-4-500 & 4 & 500 & 49.08\SI{\pm 0.5}{}\\
         \hline
         mpath-4-500 & 4 & 500 & 46.94\SI{\pm 0.2}{}\\
         \hline
         mlp-4-200 & 4 & 200 & 48.6\SI{\pm 0.2}{}\\
         \hline
         mpath-4-200 & 4 & 200 & 47.71\SI{\pm 0.2}{}\\
         \hline
         mlp-6-200 & 6 & 200 & 49.75\SI{\pm 0.3}{}\\
         \hline
         mpath-6-200 & 6 & 200 & 48.65\SI{\pm 0.1}{}\\
         \hline
    \end{tabular}
    \caption{CIFAR10 results}
    \label{tab:balance-cifar-results}
\end{table}

\begin{table}[h]
    \centering
    \begin{tabular}{c|c|c|c}
         Model & \# hidden layers & \# hidden features & Test error (\%)\\
         \hline
         mlp-4-500 & 4 & 500 & 22.57\SI{\pm 0.3}{}\\
         \hline
         mpath-4-500 & 4 & 500 & 23.27\SI{\pm 1.2}{}\\
         \hline
         mlp-4-200 & 4 & 200 & 23.7\SI{\pm 0.5}{}\\
         \hline
         mpath-4-200 & 4 & 200 & 26.1\SI{\pm 0.5}{}\\
         \hline
         mlp-6-200 & 6 & 200 & 23.1\SI{\pm 0.9}{}\\
         \hline
         mpath-6-200 & 6 & 200 & 26.83\SI{\pm 0.9}{}\\
         \hline
    \end{tabular}
    \caption{iWildCam2019 results}
    \label{tab:balance-iwild-results}
\end{table}

We test on three datasets, namely, MNIST, CIFAR10 and iWildCam2019. We use 300 epochs for the equivalent MLP training and the one-shot training of MpathNN. After convergence, we report the test error of both the equivalent MLP and MpathNN. The test error for MpathNN at this point is calculated without any divisions into independent paths by the pruning technique, i.e according to \cref{eq:balance-oneshot}, and we will refer to it by the one-shot MpathNN test error. For the sample fine-tuning phase, we use 10 evenly spaced path counts. We fine-tune the corresponding MpathNNs for 10 epochs and then report the error on the validation dataset. After that, we sample a new set of path counts that weren't used in the sample fine-tuning phase, fine-tune their corresponding MpathNNs for 30 epochs and calculate the test error. After we fit the regression model to the validation error data, we use it to predict the validation error for the new set of path counts and calculate the Pearson correlation and the MAE (Mean Absolute Error) between the test error and the predicted validation error. The Pearson correlation reflects how much the predictions can model the relative relation between the test error of different derived MpathNN models, while the MAE measures how accurate the predictions are.

For all of our experiments, we do three trials per condition, each with a different random initialization. The weights parameterizing the membrane permeabilities are initialized according to

\begin{equation}
    \Omega_i \sim \emph{U}(-1, 1)
\end{equation}

where $\emph{U}$ is the uniform distribution. We report the test error of the equivalent MLP and the one-shot MpathNN test error as an average of the three trials. We, then, randomly choose an MpathNN model from one of these three trials to conduct the sample fine-tuning phase. Pruning and sampled architecture fine-tuning will be based on this selected model. For the optimization, we use the Adam optimizer. 

From each dataset, we use a small percentage of the original training dataset as a validation dataset. The validation dataset is used for two purposes. First, calculating the validation error after each epoch during the training of the MLP and the one-shot training of the MpathNN to select the best model. Second, calculating the validation error during the sample fine-tuning phase.

\begin{figure}[h]
    \centering
    \includegraphics[width=\textwidth]{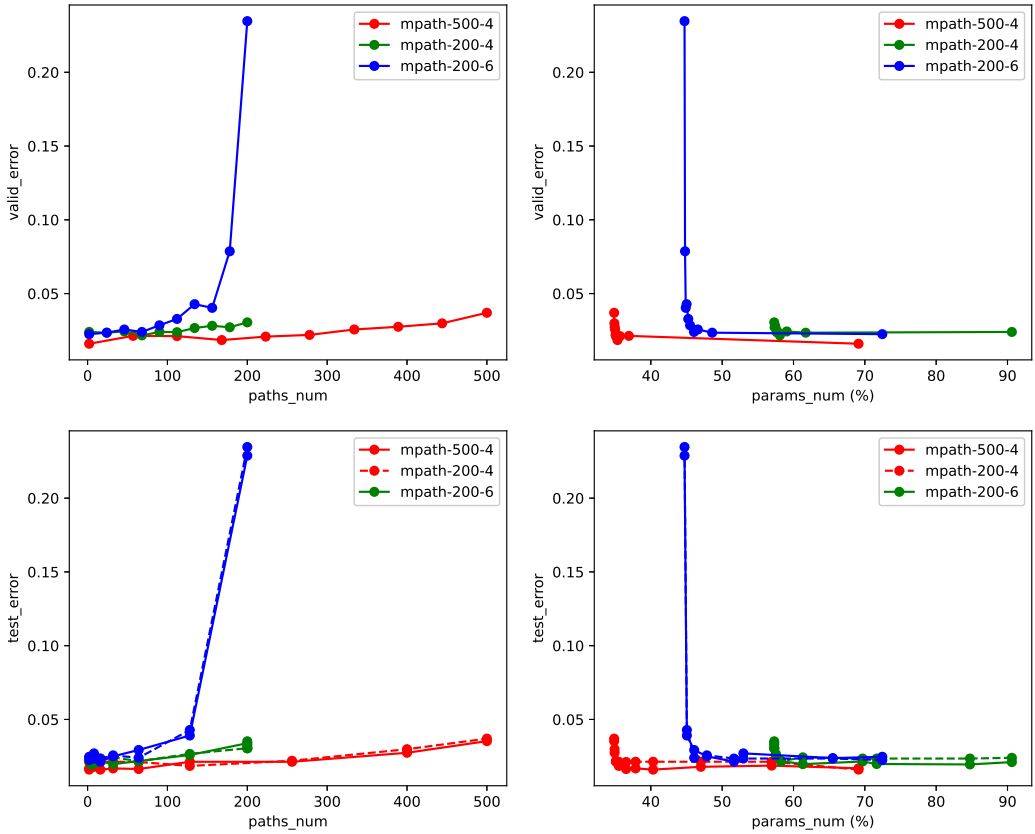}
    \caption{MNIST Sample Fine-tuning Phase. (Dotted curves represent the predicted validation error)}
    \label{fig:balance-mnist-valid}
\end{figure}

\begin{figure}[h]
    \centering
    \includegraphics[width=\textwidth]{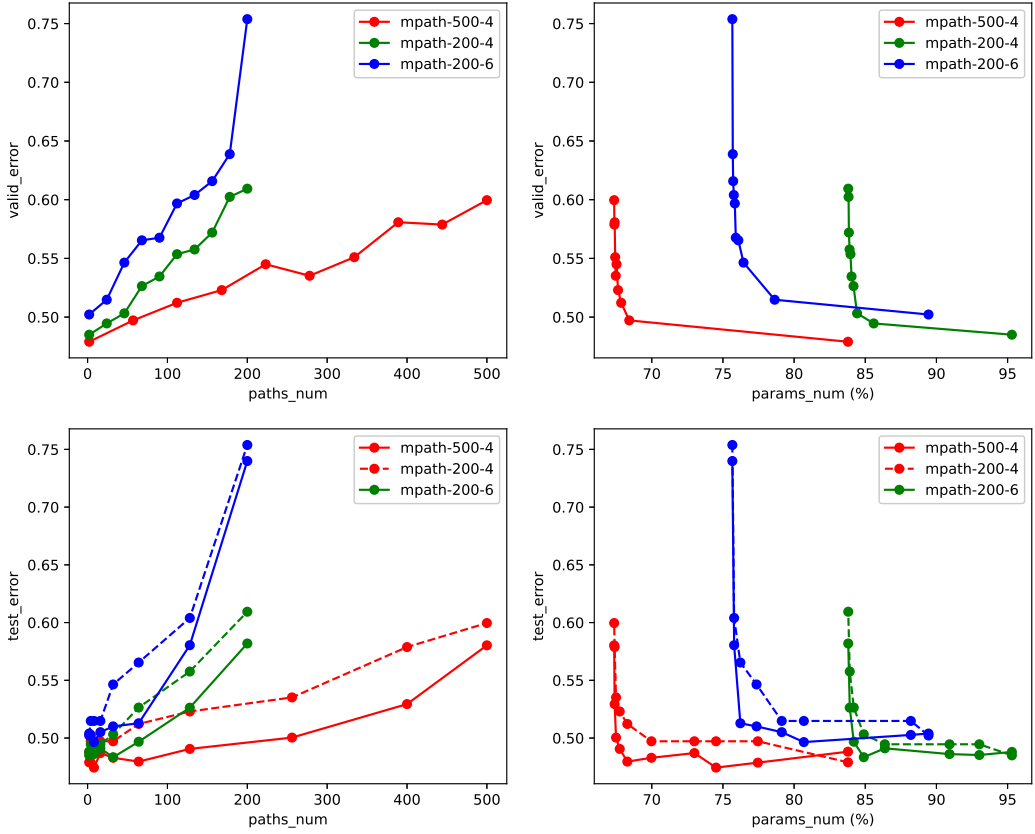}
    \caption{CIFAR10 Sample Fine-tuning Phase. (Dotted curves represent the predicted validation error)}
    \label{fig:balance-cifar-valid}
\end{figure}

\begin{figure}[h]
    \centering
    \includegraphics[width=\textwidth]{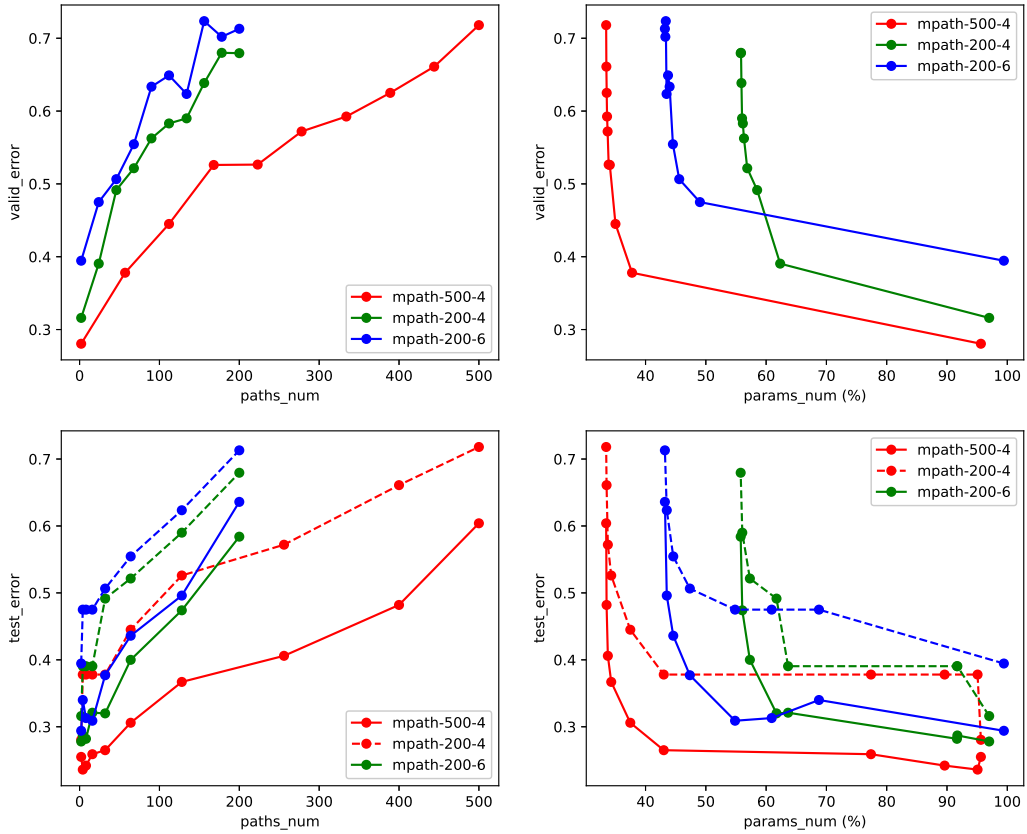}
    \caption{iWildCam2019 Sample Fine-tuning Phase. (Dotted curves represent the predicted validation error)}
    \label{fig:balance-iwild-valid}
\end{figure}

\begin{table}[h]
    \centering
    \begin{tabular}{c|c|c|c|c|c|c}
         Model & \multicolumn{3}{c|}{MAE} & \multicolumn{3}{c}{Pearson}\\
         \hline
         & MNIST & CIFAR10 & iWild & MNIST & CIFAR10 & iWild\\
         \hline
        mpath-4-500 & 0.003 & 0.02 & 0.1 & 0.91 & 0.57 & 0.92\\
        \hline
        mpath-4-200 & 0.003 & 0.02 & 0.1 & 0.91 & 0.97 & 0.95\\
        \hline
        mpath-6-200 & 0.003 & 0.02 & 0.1 & 1.0 & 0.98 & 0.98\\
        \hline
    \end{tabular}
    \caption{MAE and Pearson correlation}
    \label{tab:balance-corr-results}
\end{table}

\subsection{MNIST}

The MNIST dataset is a collection of images of handwritten digits with 10 classes representing the digits from 0 to 9. The images are grayscale with spatial dimensions of 28x28. MNIST consists of a training dataset of 60k samples and a testing dataset of 10k samples. We divide the training dataset into 90\% for training and 10\% for validation. No augmentation was used for the training data. The test dataset was used as it is. 

\subsection{CIFAR10}

The CIFAR10 dataset is a collection of colored (3-channels) images comprising 10 classes representing animals and vehicles. The images have 32x32 spatial dimensions. CIFAR10 consists of 50k training images and 10k testing images. We divide the training set into 90\% for training and 10\% for validation. No augmentation was used for the training data. The test dataset was used as it is.

\subsection{iWildCam2019}

The iWildCam2019 dataset is a collection of colored animal images captured in the wild by camera traps. The dataset is imbalanced and there is only a partial overlap between the classes found in the training and test datasets. To balance the dataset and fix the classes between training and testing, we used only the training set by splitting it into train, validation and test sets. This was done by first choosing 10 classes (deer, squirrel, rodent, fox, coyote, raccoon, skunk, cat, dog, opossum), then balancing all the classes by choosing only 1000 samples from each class. We then split the data into 70\% training, 20\% validation and 10\% test. This was done class-wise to maintain the balance. We preprocessed all the sets by converting the images to grayscale and downsampling to 23x32 spatial dimensions. No augmentation was used.

\subsection{Results}

The test performance results of the equivalent MLP models and MpathNN models after the one-shot training (i.e as per equation \cref{eq:balance-oneshot}) are listed in \cref{tab:balance-mnist-results}, \cref{tab:balance-cifar-results} and \cref{tab:balance-iwild-results}. Relative difference in performance is consistent across the three different architectures. For MNIST and iWildCam2019, MLPs perform consistently better than MpathNNs, while the reverse is true for CIFAR10.

\cref{fig:balance-mnist-valid}, \cref{fig:balance-cifar-valid} and \cref{fig:balance-iwild-valid} show different plots of the sample fine-tuning phase for the different architectures benchmarked on the three datasets. Note that while the model used for the prediction is a linear regression model, the prediction points are not aligned on a single line. This is due to the way the regression model is applied to a sequence of points as described earlier. The upper left plot of each figure shows the validation error as a function of the path count. The upper right plot shows the validation error as a function of the percentage of parameters in the corresponding MpathNN with a given division of paths. The percentage of parameters here is relative to the parameter count of the equivalent MLP, i.e $\frac{N_d}{N}\%$, where $N_d$ is the parameter count of a given division of an MpathNN and $N$ is the parameter count in an equivalent MLP. Note that, as we discussed, the mapping from a path count to its corresponding parameter count is unique under the previously described pruning technique. Hence, the path count and parameter percentage scales are unique maps of each other, however, in reverse direction, i.e a larger path count means a smaller parameter percentage.

The bottom left plot shows the actual test error as solid curves and the corresponding predicted validation error as dotted curves, both as a function of the path count. Note that we predict only the validation error of the path counts for which we are going to calculate test errors. This means that while the big dots on the dotted curve are actual predictions from the regression model, the dotted line segments are just a linear interpolation. The bottom right plot shows the same results but as a function of the percentage of parameters. The MAE and correlation between the actual test error and the predicted validation error are shown in \cref{tab:balance-corr-results}. The MAE results show an overall small deviation between the predictions and the actual errors. The Pearson correlation results as well show a consistent high correlation between the same two sets. 

\section{Discussion} \label{sec:membrane-disc}

The main idea behind our approach is using differential architectural search to rank the different divisions of an MpathNN. By subsequent pruning, a relation between parameter count and accuracy can be established and divisions can be made to meet the desired accuracy-latency requirements. Given a permissible range of latency and accuracy, a modeler can, without the need for training every potential model, meet the desired balance. Some trade-offs can't be realised due to limitations of the model and the capacity-accuracy relationship. In this case, the modeler can know beforehand and ranges can be adjusted to more realistic values.

The results of the one-shot training in \cref{tab:balance-mnist-results}, \cref{tab:balance-cifar-results} and \cref{tab:balance-iwild-results} show performance close to equivalent MLPs. The results of equivalent MLPs for MNIST and iWildCam2019 are slightly better than MpathNN. For CIFAR10, the performance of MpathNN is significantly better than MLP. This confirms that no serious under/overfitting is happening from introducing the permeability factors. 

The MAE of the sample fine-tuning phase is small and consistent, showing decent approximation for different architectures and datasets and reliability as a proxy for balancing accuracy-latency requirements. Pearson correlation also shows consistency and decent high correlation for different architectures and datasets. This supports our conclusion regarding the ability of the method to approximate the relation between latency and accuracy based on the introduced permeability factors and the pruning based on the relative magnitude of factors.

\section{Conclusion}

In this paper, we investigated using NAS to modularize a fully-connected network into sparser MpathNN architectures targeted at use cases with resource limitations. By modeling the search using a membrane permeability metaphor, we could use a one-shot model to find divisions of multiple paths that could be used efficiently to predict accuracy-latency relationships. Hence, an efficient balancing between the available resources and the desired accuracy threshold can be realised. We showed that the proposed method has good predictive power and can replace exhaustive search.

There are limitations to the described method which constitute multiple directions for future work. The simplified assumption about a limited architectural space requiring layers with the same depth can be generalized to more general architectures. More generalization could be achieved by extending to other types of NNs like convolutional neural networks. For pruning, we used weight magnitude, which has logical justification and previous usage in the literature. However, pruning is a large topic in the field and more advanced techniques may give rise to a more efficient architectural search.

\section{Acknowledgement}

This work was partially supported by a grant from Microsoft's AI for Earth program.

\bibliographystyle{plainnat}
\bibliography{main}

\end{document}